\documentclass[10pt,twocolumn,letterpaper]{article}

\usepackage[review]{cvpr}

\usepackage{float}
\usepackage{multirow}
\usepackage{float}      
\usepackage{afterpage}
\usepackage{scalerel,xparse}
\usepackage{capt-of,etoolbox}%

\newcommand{\ignorethis}[1]{}

\definecolor{cvprblue}{rgb}{0.21,0.49,0.74}
\usepackage[pagebackref,breaklinks,colorlinks,allcolors=cvprblue]{hyperref}

\def\confName{CVPR}
\def\confYear{2025}

\title{ScribbleLight: Single Image Indoor Relighting with Scribbles}

\author{Jun Myeong Choi$^{1}$, Annie Wang$^{1}$, Pieter Peers$^{2}$, Anand Bhattad$^{3}$, Roni Sengupta$^{1}$ \vspace{1mm}\\
$^{1}$University of North Carolina at Chapel Hill, $^{2}$College of William $\&$ Mary, \\$^{3}$Toyota Technological Institute at Chicago
}

\begin{document}
\makeatletter
\g@addto@macro\@maketitle{
	\begin{figure}[H]
	\scriptsize
		\setlength{\linewidth}{\textwidth}
		\setlength{\hsize}{\textwidth}
            \vspace{-4mm}
  \centering
  \footnotesize
  \setlength\tabcolsep{0.2pt}
  \renewcommand{\arraystretch}{0.1}
\includegraphics[width=\linewidth]{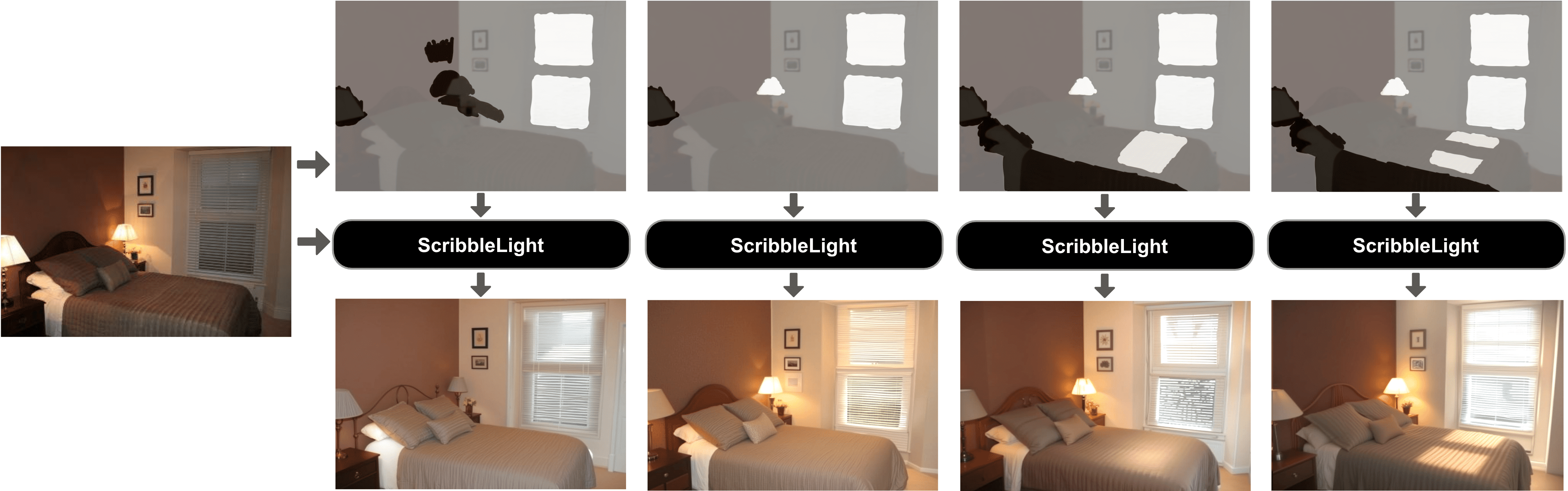} 
\vspace{-5pt}

\caption{We introduce ScribbleLight, a generative model designed for indoor scene relighting from a single RGB image, that allows users to iteratively refine lighting effects in a photo using simple scribble annotations. In the first scribble, natural light from the window changes the scene from nighttime to daytime by turning off both lamps and brightening the area near the window (left side) and casting a soft glow on the bed. In the second scribble, only the right lamp is turned on. In the third scribble, the natural light intensifies, increasing the gloss on the bed’s surface. In the final scribble, adding the split in the light cast from the window causes the angle of the incoming light to change, creating a strong contrast with the warm glow of the bedside lamp.}

        \label{fig:teaser}
        \vspace{-2pt}
	\end{figure}
}
\makeatother    
\maketitle

\vspace{-1em}
\begin{abstract}
Image-based relighting of indoor rooms creates an immersive virtual understanding of the space, which is useful for interior design, virtual staging, and real estate. Relighting indoor rooms from a single image is especially challenging due to complex illumination interactions between multiple lights and cluttered objects featuring a large variety in geometrical and material complexity. Recently, generative models have been successfully applied to image-based relighting conditioned on a target image or a latent code, albeit without detailed local lighting control.
In this paper, we introduce ScribbleLight, a generative model that supports local fine-grained control of lighting effects through scribbles that describe \emph{changes} in lighting. Our key technical novelty is an Albedo-conditioned Stable Image Diffusion model that preserves the intrinsic color and texture of the original image after relighting and an encoder-decoder-based ControlNet architecture that enables geometry-preserving lighting effects with normal map and scribble annotations.  We demonstrate ScribbleLight's ability to create different lighting effects (\eg, turning lights on/off, adding highlights, cast shadows, or indirect lighting from unseen lights) from sparse scribble annotations. For more information, please visit our website: \url{https://chedgekorea.github.io/ScribbleLight/}.

\end{abstract}    
\section{Introduction}
\label{sec:intro}

In today’s digital age, the ability to control and visualize lighting in indoor spaces is crucial, especially in downstream applications such as real estate, virtual staging, and interior design. Traditional static images of indoor scenes often fail to show how a room would look under different lighting conditions, missing key elements that affect the space's aesthetics and mood. Scene relighting techniques offer a solution by enabling dynamic lighting adjustments within an image, letting users envision how a room transforms under various lighting scenarios, such as natural sunlight or with different lamps in the room turned on. The ability to relight provides a deeper and immersive understanding of the space, without needing a physical visit.

Indoor relighting presents unique challenges compared to outdoor settings, where a single, predictable light source (\ie, Sun) yields consistent and directional shadows; the resulting limited model space has been successfully exploited in prior outdoor relighting methods~\cite{outdoor_1, outdoor_2, outdoor_3, outdoor_4, kocsis2024lightit}.  In contrast, indoor environments involve multiple light sources, such as ceiling lights, lamps, window-filtered daylight, as well as invisible light sources, each with unique characteristics in intensity, direction, and diffusion. These overlapping light sources create intricate, soft, layered shadows which create a challenging relighting environment. Indoor scenes are also composed of multiple objects made of different materials (\eg, furniture and doors), making indoor  relighting more difficult than single-object relighting.

Despite recent advances, existing state-of-the-art indoor 3D relighting methods still require significant effort in dense scene capture~\cite{indoor_3d_1,indoor_3d_2,indoor_3d_3,indoor_3d_4,indoor_3d_5}. Conversely, implicit relighting methods that use a latent space~\cite{StyLitGAN} or a reference image \cite{zhang2024latentintrinsicsemergetraining} to control lighting can only induce coarse global lighting changes and cannot control local details. Motivated by the success of using scribbles to guide various image manipulation tasks \cite{scribble_color,scribble_inpaint1,scribble_inpaint2,scribble_relit}, we propose ScribbleLight, where a user can intuitively and directly control relative lighting adjustments in an image with scribbles. ScribbleLight enables the user to iteratively indicate areas in the image they wish to brighten or darken with scribbles, from coarse to fine-grained, based on their preferences. Our generative model can relight the image from the scribble input to create various illumination effects - turning lights on and off, adding cast shadows, highlights, inter-reflections, etc. Our method bridges the gap between technical complexity and creative flexibility, offering a streamlined, user-friendly way to achieve professional-quality lighting adjustments in intricate indoor scenes.

Scribbles only offer high-level guidance. Consequently, to resolve the inherent guidance-ambiguities, we exploit the general image prior embedded in large pretrained generative diffusion model (\ie, Stable Diffusion v2~\cite{rombach2022stablediffusion}) and control the lighting effects with ControlNet~\cite{zhang2023adding}.
However, a naive implementation fails to preserve the color and texture, \ie, intrinsic albedo, of the original image in the relit image. We therefore introduce an Albedo-conditioned Stable Image Diffusion model that generates realistic images conditioned on the intrinsic albedo of the scene. 
To support large lighting changes and to improve robustness with respect to an imperfect albedo (predicted by an Intrinsic Image Decomposition~\cite{careaga2024iid}) we inject uncertainty in the training process by adding noise to the (albedo) condition, thereby reducing dependency on the exact content of the albedo and forcing the diffusion model to rely more on the embedded image prior.  We add lighting control during the albedo-conditioned diffusion process via a ScribbleLight ControlNet, where the control signal is the latent embedding of the scribbles and normals obtained from an encoder-decoder network that reconstructs the normal and \emph{shading} map from the input. The decoder's ability to predict the intended shading (from scribles) and reconstruct the normals improves the likelihood that the latent code includes all the necessary information for relighting.

ScribbleLight enables flexible lighting control while retaining the intricate color and texture details of the original scene, overcoming challenges that arise due to the sparsity of scribbles. As no other prior indoor relighting method performs scribble-driven single-image relighting of indoor rooms, we compare our approach to baselines derived from existing approaches~\cite{kocsis2024lightit,zeng2024rgbx} that also use Stable Diffusion and ControlNet for relighting. Quantitative and qualitative evaluations show that our method significantly outperforms the baseline methods, which often fail to preserve the albedo of the input image and control local lighting details. We also perform an extensive ablation study to demonstrate the effectiveness of ScribbleLight. 

\begin{figure*}[h]
  \centering
  \includegraphics[width=.9\linewidth]{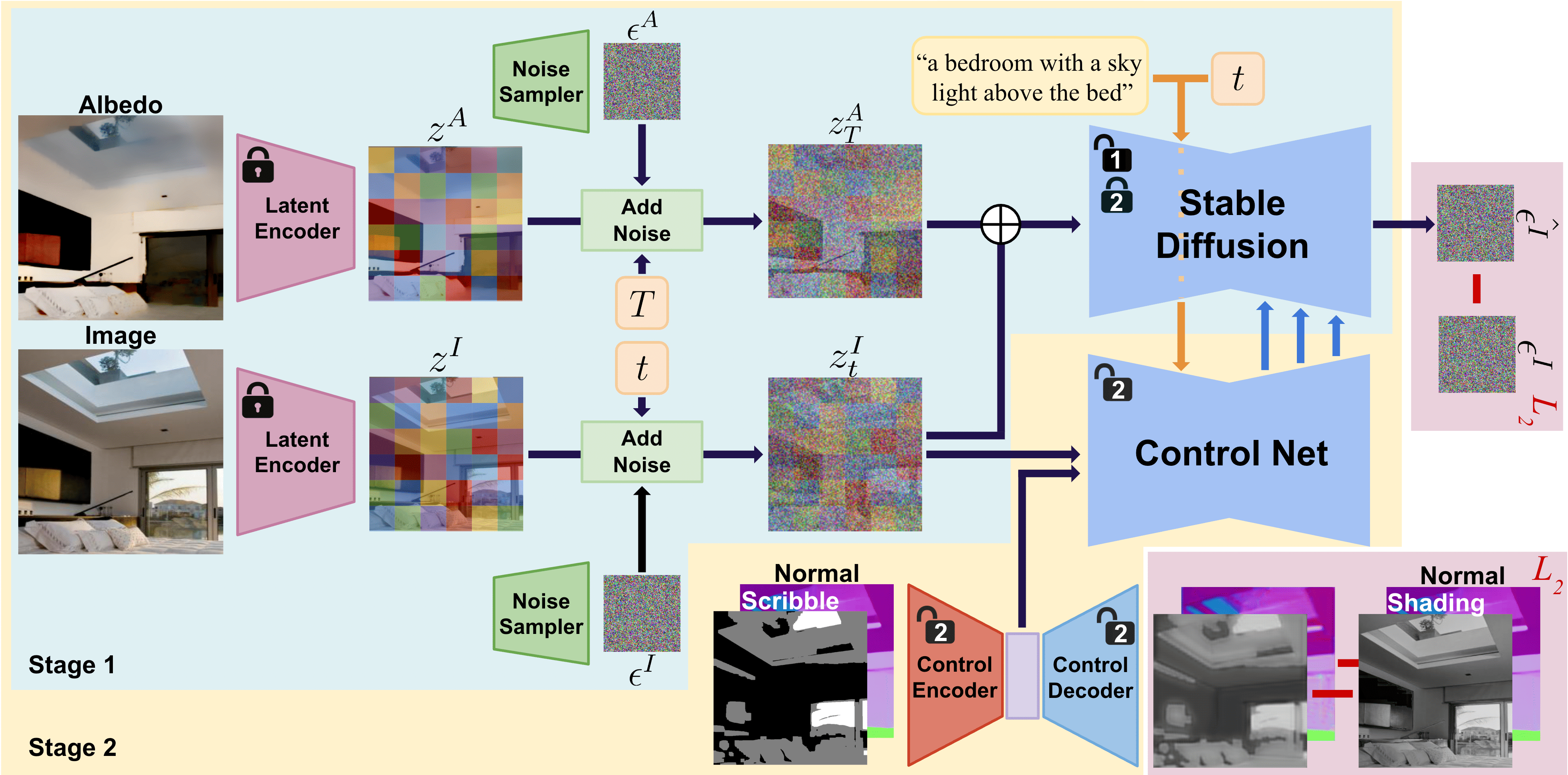}
  \caption{ScribbleLight consists of an Albedo-conditioned Stable image Diffusion model (trained in Stage 1), and a ControlNet (trained in Stage 2) that guides the albedo-conditioned diffusion model for relighting through a latent encoding of the scribbles and normals.  To regularize the latent encoding, we jointly train a decoder that predicts the target shading (and normals) from the scribbles (and normals).}
  \label{fig:model_architecture}
  \vspace{-1.5em}
\end{figure*}

\section{Related work}
\label{sec:related_work}

\noindent \textbf{Image-based relighting} aims to alter the lighting in photographs post-capture.  Specialized methods have been proposed for relighting isolated objects~\cite{object_2, object_6, object_7, object_8, object_9, object_10,object_11,object_12,object_13}, human portraits~\cite{portrait_1, portrait_2, portrait_3, portrait_4, portrait_5, portrait_6, scribble_relit, choi2024personalizedvideorelightingathome}, human bodies~\cite{human_1, human_2, human_3}, outdoor scenes~\cite{kocsis2024lightit, outdoor_1, outdoor_2, outdoor_3, outdoor_4}, and indoor scenes~\cite{jundan2024intrinsicdiffusion,kocsis2024iidiffusion,zeng2024rgbx,zhang2024latentintrinsicsemergetraining,StyLitGAN}.  The focus of our work, indoor scene relighting, is especially challenging due to the mixture of natural and artificial light sources, occlusions, and intricate light interactions in a cluttered scene creating cast shadows, strong highlights, and inter-reflections.

Image-based relighting research has explored different lighting representations to control illumination in the rendered image. Explicit lighting representations such as shadow-maps~\cite{jundan2024intrinsicdiffusion}, spherical Gaussians~\cite{kocsis2024iidiffusion}, or irradiance fields~\cite{zeng2024rgbx} directly specify the lighting, thereby offering the user only indirect control on the \emph{effects} of lighting in the scene (\ie, the goal of the user). Alternatively,  Zhang~\etal~\cite{zhang2024latentintrinsicsemergetraining} and Bhattad~\etal~\cite{StyLitGAN} use implicit lighting representations instead and navigate the latent space to control lighting effects.
However, latent space editing only offers coarse global control and cannot control local details, making it difficult for the user to achieve their exact goal. In this paper, we use user-friendly scribbles to enable more fine-grained control of the lighting effects. 

\noindent \textbf{Image manipulation using scribbles}
offers an intuitive interface for specifying a user's intent. Scribbles have been used as a guide in a wide variety of tasks such as: segmentation~\cite{scribble_seg1, scribble_seg2, scribble_seg3, scribble_seg4}, image generation~\cite{scribble_gen1, scribble_gen2, scribble_gen4,pix2pix,scribble_gen3}, image  editing~\cite{scribble_edit1,scribble_edit2,scribble_edit3,scribble_edit4,scribble_edit5,scribble_edit6}, inpainting~\cite{scribble_inpaint1, scribble_inpaint2}, retrieval\cite{scribble_retrieval1,scribble_retrieval2} , and colorization~\cite{scribble_color}.
Similar to us, Mei~\etal~\cite{scribble_relit} use scribbles to control relighting of human portraits.  However, the geometrical complexity and material and lighting variations in indoor scenes make scribble-based relighting guidance more challenging.

\noindent \textbf{Intrinsic image decomposition} (IID) is a fundamental problem in computer vision that aims to separate an image into an illumination-dependent component (\ie, shading) and an illumination-independent component (\ie, reflectance or albedo).  Early IID methods rely on heuristics based on physical properties or empirical observations~\cite{grosse2009ground, land1971lightness, garces2012intrinsic, chen2013simple, barron2013intrinsic, barron2014shape}, often limited to Lambertian or simple scenes. Recent IID methods leverage machine learning trained on synthetic data~\cite{neuralSengupta19,li2020inverse, zhu2022learning, baslamisli2018joint, careaga2023intrinsic, das2022pie, janner2017self, li2018cgintrinsics, Li2018BigTime, kocsis2024iidiffusion} to support more complex scenes and non-Lambertian reflectance.
Relighting is a common downstream task of IID, which is achieved by changing the illumination-dependent component and recompositing the intrinsic components. However, to achieve a plausible relit result, the modified illumination-dependent component has to contain all the details, making it a cumbersome and error-prone interface for relighting.  In contrast, scribbles do not require the user to provide pixel-precise shading details, thereby providing a more convenient and user-friendly control interface.

\section{Method}
\label{sec:method}

We aim to generate plausible relit images of indoor scenes from a single photograph and guided by user-provided scribbles. Instead of viewing the guidance as target pixel or shading values, we instead view scribbles as a way for the user to indicate which areas need to be brightened (\eg, turning on a light) and which areas need to be darkened (\eg, adding a cast shadow). Therefore, we employ a binary scribble with `1' indicating brightening, and `0' indicating darkening. Unlabeled areas are left to the relighting model to determine the most plausible action.

Inspired by recent successes in using generative diffusion models for relighting tasks~\cite{portrait_2,Bashkirova:2023:Lasagna,Ding:2023:DLP,object_6,kocsis2024lightit}, we introduce ScribbleLight, a ControlNet-based single-image relighting solution for guiding an albedo-conditioned diffusion model (\cref{sec:stage1}) with scribbles and normals (\cref{sec:stage2}). Our training pipeline involves two distinct stages: first, fine-tuning an albedo-conditioned Stable Diffusion model, followed by separate training of the scribble-guided ControlNet. \Cref{fig:model_architecture} summarizes our pipeline. 

\subsection{Albedo-conditioned Image Diffusion}
\label{sec:stage1}
A key observation underpinning ScribbleLight is that relighting should preserve the underlying albedo intrinsic, i.e. color and texture, of the input photograph.   While Stable Diffusion v2~\cite{rombach2022stablediffusion} trained on LAION-5B~\cite{schuhmann2022laion5b} provides a strong image prior, it lacks a strong constraint on the underlying albedo intrinsic.   Therefore, we refine Stable Diffusion to produce an image~\textbf{I} conditioned on an additional albedo image~\textbf{A}. This will help the diffusion model to maintain the input image's color and texture information during relighting. Concretely, we employ the pre-trained Latent VAE Encoder $\mathcal{E}^L$ to encode both the image $\textbf{I}$ and the corresponding albedo $\textbf{A}$ into the latent space, \ie, Image Latent ($z^{I} = \mathcal{E}^L(I)$) and Albedo Latent ($z^{A} = \mathcal{E}^L(A)$). 

We follow Stable Diffusion's training process by adding $\epsilon^{I}_{t}$ noise to the image latents $z^{I}$ for a randomly sampled time step $t \in \{ 1,...,T \}$ and learning to denoise to $z^{I}_{t-1}$.  However, directly conditioning the diffusion process on the albedo image poses two problems. First, because the scribbles do not encode any spatially varying intensity changes, the diffusion model tends to produce relit regions with little variation. Second, any errors present in the albedo map are included in the diffusion process yielding visually noticeable artifacts.  We resolve both problems by also adding a fixed amount of noise $\epsilon^{A}_T$ to the albedo latents $z^{A}$ (\ie, introduce uncertainty) to preserve the fundamental color and structure of the scene while providing enough uncertainty to render different lighting conditions.  In contrast to the image noise which varies per time step $t$, $\epsilon^{A}_T$ remains fixed at the level of $T=200$, an optimal value we observed empirically. Next, we concatenate $z^{I}_{t}$ and $z^{A}_{T}$ into a single input, $z_t$, along the feature dimension. This results in a doubling of the input channels of the latent denoising U-net in Stable Diffusion, and we zero-initialize the additional convolution weights. Finally, we train the albedo-conditioned Stable Diffusion model ($\theta^S$) with text prompt $p$ using the following modified loss function:
\begin{equation}
  \mathcal{L} = \mathbb{E}_{z^{I}_{t},z^{A}_{T}, \epsilon^{I}\sim\mathcal{N}(0,1),t,p}	\left[  \left\| \epsilon - \epsilon_{\theta^S}(z^{I}_{t},z^{A}_{T}, t, p) \right\|^2_2\right].
  \label{eq:loss_stablediffusion}
\end{equation}

\subsection{ScribbleLight ControlNet}
\label{sec:stage2}
We employ ControlNet~\cite{zhang2023adding} to guide the albedo-conditioned image diffusion using a user-provided scribble map \textbf{M} and normal map \textbf{N} to generate a relit image. We first concatenate and encode the scribble map \textbf{M} and normals \textbf{N} into a lighting feature map ($f = \mathcal{E}^C([\textbf{M}, \textbf{N}])$) using a learnable control encoder $\mathcal{E}^C$.  To regularize the control encoder, we introduce an additional control decoder $\mathcal{D}^C$ that recovers the normal map \textbf{N} and predicts a monochromatic intrinsic shading component $\textbf{S}_{mono}$ from the lighting feature map:
\begin{equation}
\begin{split}
  \mathcal{L}_{D} = \left\| \mathcal{D}^C(\mathcal{E}^C(\textbf{M},\textbf{N})) - (\textbf{S}_{mono}, \textbf{N}) \right\|^2_2.
\end{split}
  \label{eq:loss_condition}
\end{equation}
The Control Encoder-Decoder architecture ensures that the latent lighting features contain the scene geometry and shading information necessary for relighting.

The ControlNet takes as input the lighting feature map $f$, image latent code $z^{I}_{t}$ at time-step $t$, and the text prompt $p$.  Because our ControlNet is not conditioned on the albedo \textbf{A}, we instead initialize it with the original prompt-conditioned Stable Diffusion v2 weights, and train it jointly with the Control Encoder-Decoder using the following loss:
\begin{equation}
\begin{split}
  \mathcal{L} = \mathcal{L}_{D} +
  \mathbb{E}_{z_{t}, f, \epsilon^{I}\sim\mathcal{N}(0,1),t,p}	\left[  \left\| \epsilon - \epsilon_{\theta^C}(z_{t}, f, t, p) \right\|^2_2\right] .
\end{split}
  \label{eq:loss_controlnet}
\end{equation}

\subsection{Training Data and Scribble Generation}
\label{sec:scribbleGeneration}
Existing large-scale indoor relighting datasets such as InteriorVerse~\cite{zhu2022learning} and OpenRooms~\cite{li2021openrooms} consist of synthetic scenes rendered from different views under two or more lighting conditions.  There is a significant domain gap between these datasets and real photographs. To reduce the domain gap we opt to train ScribbleLight on the real indoor images from LSUN Bedrooms~\cite{yu2015lsun}. 

To train the albedo-conditioned image diffusion model (\cref{sec:stage1}), we require an albedo map $\textbf{A}$ for each image $\textbf{I}$ from the training set. We compute $\textbf{A}$ using a state-of-the-art IID~\cite{careaga2024iid}.  We also require a corresponding text prompt $\textbf{p}$ that we generate using BLIP-2~\cite{li2023blip2} from the image \textbf{I}. 

Training the ControlNet (\cref{sec:stage2}) requires the normals \textbf{N} and scribbles \textbf{M}. Additionally, to train the Control-decoder, we also require a corresponding monochromatic shading image $\textbf{S}_{mono}$.  The normal $\textbf{N}$ are computed using DSINE~\cite{bae2024dsine} and the shading $\textbf{S}_{mono}$ are provided by an IID method~\cite{careaga2023intrinsic}.  We generate the scribbles $\textbf{M}$ automatically from the shading $\textbf{S}_{mono}$ by setting
$\textbf{M}(x) = 1$ when $\textbf{I}(x) > \mu + \sigma$, \textbf{M}(x) = 0, when $\textbf{I}(x) < \mu - \sigma$, and $\textbf{M}(x) = 0.5$ otherwise, where $\mu$ and $\sigma$ are the mean and standard deviation of the pixel intensity distribution in the training data.  However, the thresholded scribbles exhibit edges that strongly align with the content in the input image $\textbf{I}$, something unlikely to happen with user-drawn scribbles. Therefore, we  perform an additional dilation and erosion with a kernel size randomly sampled between $3$ and $19$.

\begin{figure*}[h]
  \includegraphics[width=\linewidth]{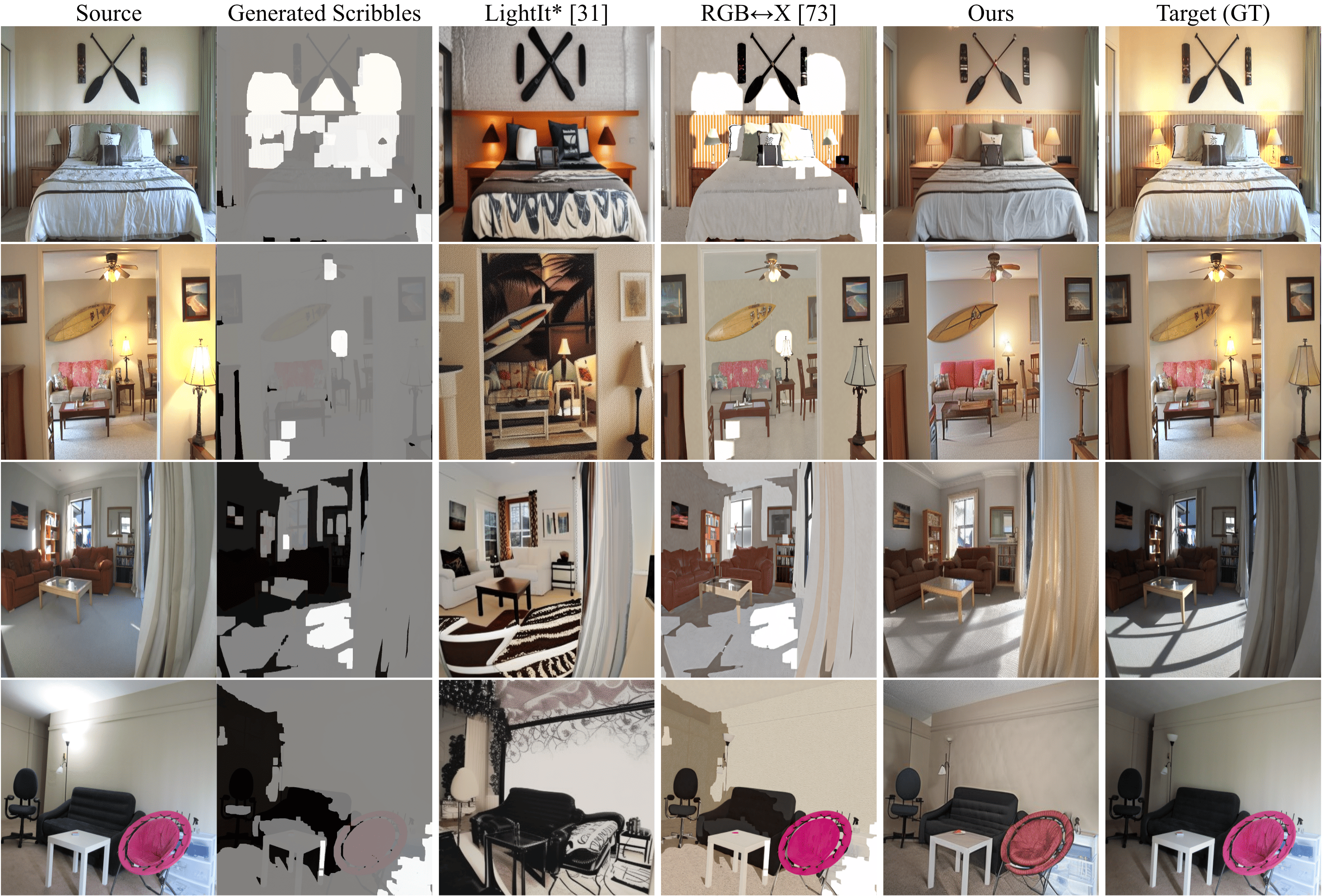}
  \caption{Qualitative comparison of relighting quality between LightIt*~\cite{kocsis2024lightit}, RGB$\leftrightarrow$X~\cite{zeng2024rgbx} and ScribbleLight (Ours) with auto-generated scribbles given a target (GT) image.}
  \label{fig:auto_scrib}
  \vspace{-1em}
\end{figure*}

\section{Experiment}
\label{sec:experiment}

\subsection{Evaluation Framework}
\label{sec:eval_framework}

\noindent \textbf{Dataset.} 
We use a subset of $100$K images from the LSUN Bedrooms dataset~\cite{yu2015lsun} as our training data, and we select $206$ pairs of indoor room images, each pair with the same environment but two different lighting conditions, from the BigTime time-lapse dataset~\cite{Li2018BigTime} for testing.
We generate automatic scribble annotations from the IID generated shading map following~\cref{sec:scribbleGeneration}.  We also manually create hand-drawn scribble annotations for a small selection of images from LSUN Bedrooms and publicly available internet images. To differentiate between both, we denote the former as \emph{`auto-generated scribbles'} and the latter as \emph{`user scribbles'}.  For illustration purposes, we display the scribble annotations overlaid over the source image; we use clean scribbles for computations.

\noindent \textbf{Baselines.} To our knowledge no prior method can guide indoor room relighting with scribble annotations.  Therefore, we adapt two existing image-based relighting algorithms as a comparison baseline. As a first baseline method, we retrain \emph{LightIt}~\cite{kocsis2024lightit} on our training dataset, and directly pass the auto-generated scribble annotations as input to the control-encoder instead of the monochromatic shading map; we denote our retrained version as \emph{`LightIt*'}. We employ \emph{RGB$\leftrightarrow$X}~\cite{zeng2024rgbx} without retraining as our second baseline method; we first generate the intrinsic components (normal, albedo, roughness, and metallicity)  using RGB$\rightarrow$X, and then recompose the image using X$\rightarrow$RGB with the shading map replaced by the scribble annotations.

\noindent \textbf{Metrics.}
We employ four error metrics to assess performance: RMSE, PSNR, SSIM, and LPIPS~\cite{perceptual_loss}. Lower RMSE and higher PSNR indicate better per-pixel similarity to the reference. SSIM (higher is better) assesses structural similarity, while LPIPS (lower is better) evaluates intrinsic feature similarity. We compute both the average and best values for each metric over $5$ replicates (\ie, different diffusion seeds). Note that raw metric values do not necessarily capture the full relighting quality of the different outputs, but rather only measure the error with respect to a single reference image. For example, a low PSNR does not necessarily indicate poor performance since a relit result can be realistic while at the same time differ from the reference.

\subsection{Evaluation with Auto-Generated Scribbles}
\label{sec:auto-result}

We first quantitatively compare the relighting quality with auto-generated scribble annotations over the test set. The results are summarized  in~\cref{tab:main} and show that ScribbleLight significantly outperforms the baseline methods across all metrics. 
\cref{fig:auto_scrib} qualitatively confirms the superior performance.  Even though the rough scribble annotations do not include detailed shadow and shading information,  ScribbleLight is able to produce a relighting result similar to the target image. In contrast, the RGB$\leftrightarrow$X results appear to overlay the coarse scribble annotation yielding an unrealistic result. LightIt* is able to interpret the scribble annotations correctly, but fails to preserve the albedo of the source image, resulting in lighting that differs significantly from the target. In the appendix, we demonstrate that even with $\textbf{S}_{mono}$ (instead of scribbles), ScribbleLight outperforms RGB$\leftrightarrow$X and LightIt*.

\begin{table}
  \centering
  \resizebox{\linewidth}{!}{%
  \begin{tabular}{lcccc}
    \toprule
     & {\bfseries{RMSE} $\downarrow$} & {\bfseries{PSNR} $\uparrow$} & {\bfseries SSIM $\uparrow$} & {\bfseries LPIPS $\downarrow$} \\
    \midrule
    LightIt* 
    & 0.341(0.302) & 9.61(10.65) & 0.232(0.332) & 0.564(0.518)  \\
    RGB$\leftrightarrow$X 
    & 0.269(0.251) & 12.47(12.99) & 0.416(0.437) & 0.439(0.425)  \\
    Ours  
    & \textbf{0.206}(\textbf{0.190}) & \textbf{14.29}(\textbf{15.01}) & \textbf{0.436}(\textbf{0.504}) & \textbf{0.394}(\textbf{0.370})  \\
  \bottomrule
\end{tabular}%
}
\caption{Quantitative comparison of relighting accuracy between our ScribbleLight, RGB$\leftrightarrow$X~\cite{zeng2024rgbx} and LightIt*~\cite{kocsis2024lightit}.  We compute the mean(best) errors with respect to a target image. 
}
\label{tab:main}
\end{table}

\begin{figure*}[h]
  \includegraphics[width=\linewidth]{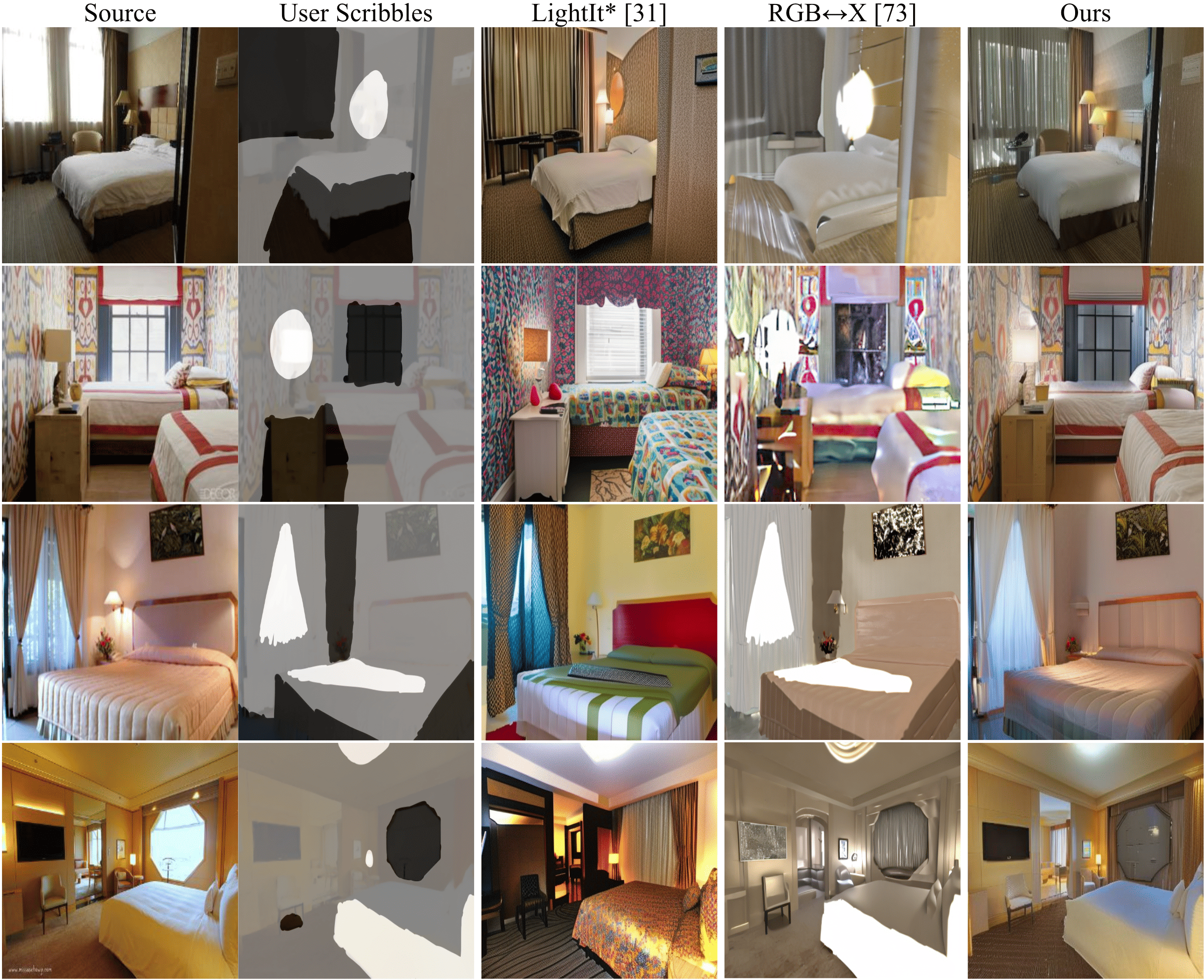}
  \caption{Qualitative comparison of relighting quality of LightIt*~\cite{kocsis2024lightit}, RGB$\leftrightarrow$X~\cite{zeng2024rgbx} and ScribbleLight (Ours) with user-provided hand-drawn scribbles.}
  \label{fig:user_scrib}
\end{figure*}

\begin{figure*}[h]
  \includegraphics[width=\linewidth]{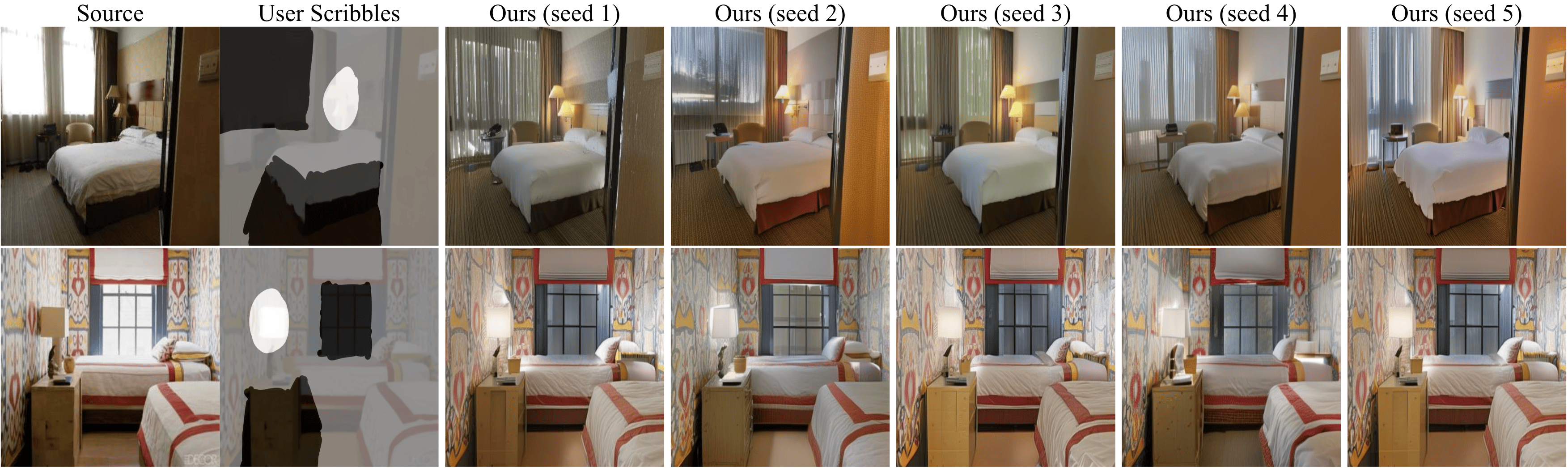}
  \caption{Our method consistently follows the scribble, even with different random seeds, allowing users to select their preferred result.}
  \label{fig:seed}
\end{figure*}

\begin{figure*}[h]
  \includegraphics[width=\linewidth]{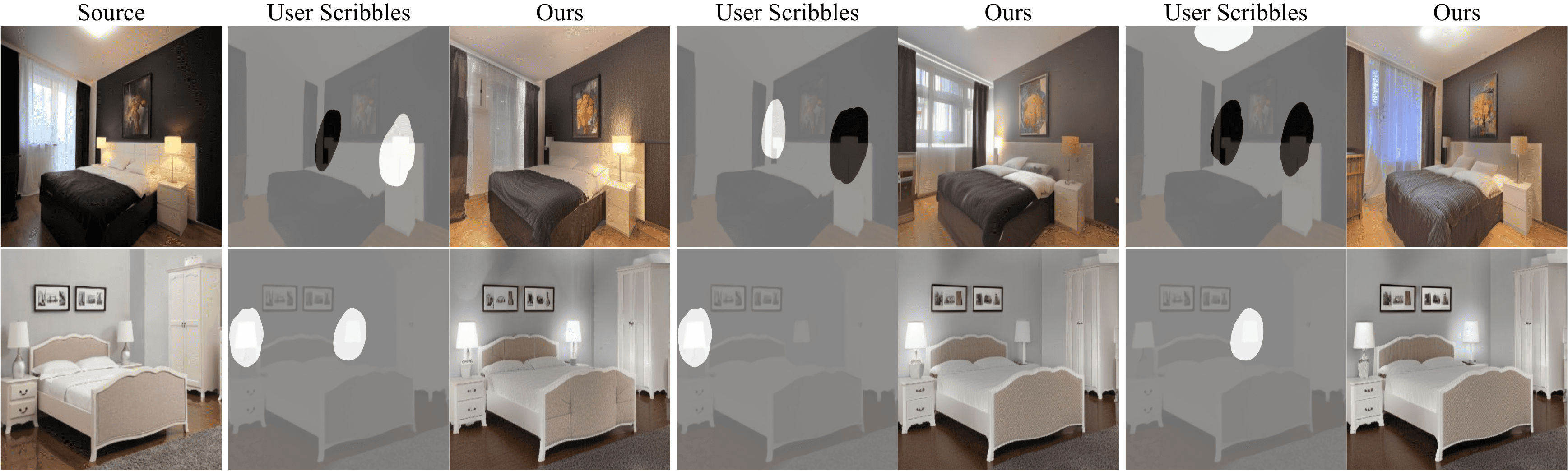}
  \vspace{-1.5em}
  \caption{Demonstration of ScribbleLight's ability to generate different plausible relit images by turning on and off different lights while maintaining the intrinsics of the input photograph.}
  \label{fig:on_off}
\end{figure*}

\begin{figure*}[h]
  \includegraphics[width=\linewidth]{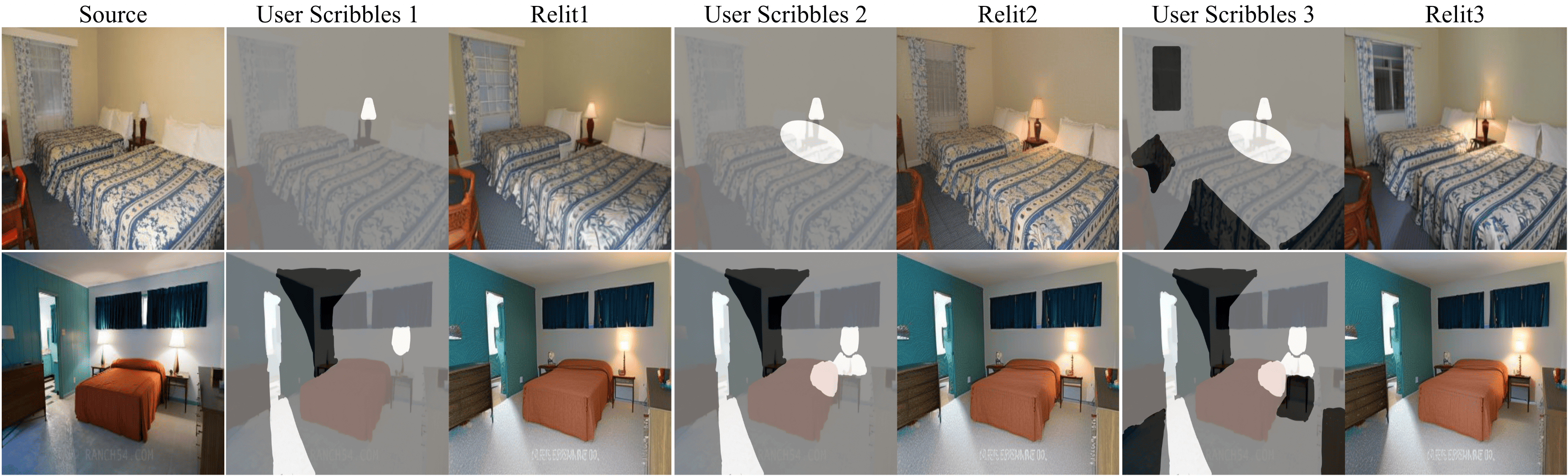}
  \vspace{-1.5em}
  \caption{Minor changes to the scribbles yield proportional changes in the relit results, enabling a user to iteratively refine the scribbles to achieve the desired results.}
  \label{fig:iterative}
  \vspace{-0.5em}
\end{figure*}

\subsection{Evaluation with User Scribbles}
\label{sec:user-result}
We perform a number of qualitative tests to demonstrate that ScribbleLight also performs well on hand-drawn user scribbles. \Cref{fig:user_scrib} shows examples of turning lights on and off (rows 1 \& 2), as well as to brighten or darken the incoming light from outside (rows 3 \& 4) using scribble annotations. In addition, we also edit shadows by marking areas for darkening (rows 1-4).  While ScribbleLight produces plausible results, RGB$\leftrightarrow$X and LightIt* fail to create realistic relighting results, producing similar artifacts as seen in \cref{sec:auto-result}.  We observe that even when the scribbles are physically inconsistent (\eg, the bright annotation on the side of the bed (row 4) is unlikely to be cast from a small light on the ceiling), ScribbleLight  manages to create a physically plausible result by imagining a light source outside the image with appropriate shading effects such as the gradually decreasing lighting intensity on the side of the bed. 

A key advantage of ScribbleLight is that it can generate different replicates from the same inputs by simply using a different diffusion seed (\cref{fig:seed}), allowing the user to pick their preferred solution and empowering the user to modify the lighting variations unconstrained by the scribbles. We observe that despite the changes in random seed, ScribbleLight reliably follows the scribble guidance and retains the input image's intrinsic color and texture.

Despite the simplicity of the scribble annotations, ScribbleLight can still exert precise control over lighting conditions/effects. \cref{fig:on_off} shows relit results of a source image with two lights turned on (row 1) or off (row 2), which can be separately toggled on or off with the appropriate scribbles. In both cases, ScribbleLight produces realistic soft highlights on the nearby walls from the lamps, even though this is not specified in the scribble annotations.

Finally, we demonstrate in~\cref{fig:teaser} and~\cref{fig:iterative} how users can progressively refine the scribble annotations to improve the relit results, yielding a flexible and intuitive indoor scene relighting experience. For instance, in \cref{fig:iterative} row 1, starting from a source image with the lamp turned off, we first turn on the lamp (user scribble 1), then add ambient lighting around the lit lamp (user scribble 2) creating a soft glow effect, and finally enhance realism by adding cast shadows near the bed and darkening the window. 

\begin{table}
    \centering
    \includegraphics[width=\linewidth]{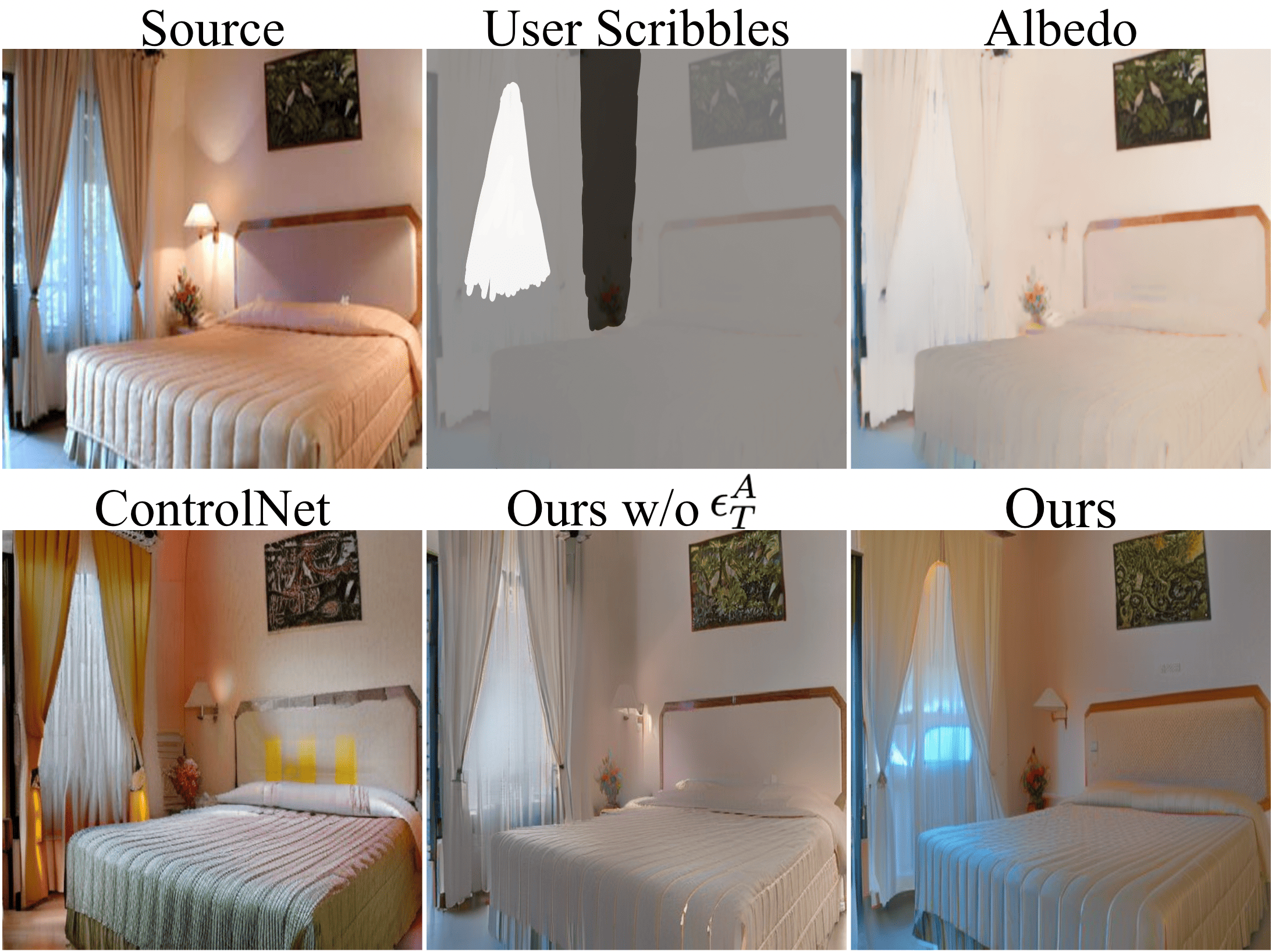} \\
      \resizebox{0.85\linewidth}{!}{%
    \begin{tabular}{cccccc}
        \toprule
         Albedo & Add $\epsilon^{A}_T$ & \bfseries{RMSE $\downarrow$} & \bfseries{PSNR $\uparrow$} & \bfseries{LPIPS $\downarrow$} \\
        \midrule
        ControlNet & \textendash & 0.2305 & 13.19 & 0.4839\\
        StableDiff & \textendash & 0.2082 & 14.07 & 0.4193\\
        StableDiff & \checkmark & \textbf{0.2059} & \textbf{14.29} & \textbf{0.3942} \\
        \bottomrule
    \end{tabular}
    }
    \caption{Design of Albedo-conditioned Stable Image Diffusion. We show that adding noise to the albedo latent improves albedo preservation and generation of realistic lighting effects.}
    \label{tab:ablation_noisy}
\end{table}

\begin{table}
    \centering
    \includegraphics[width=\linewidth]{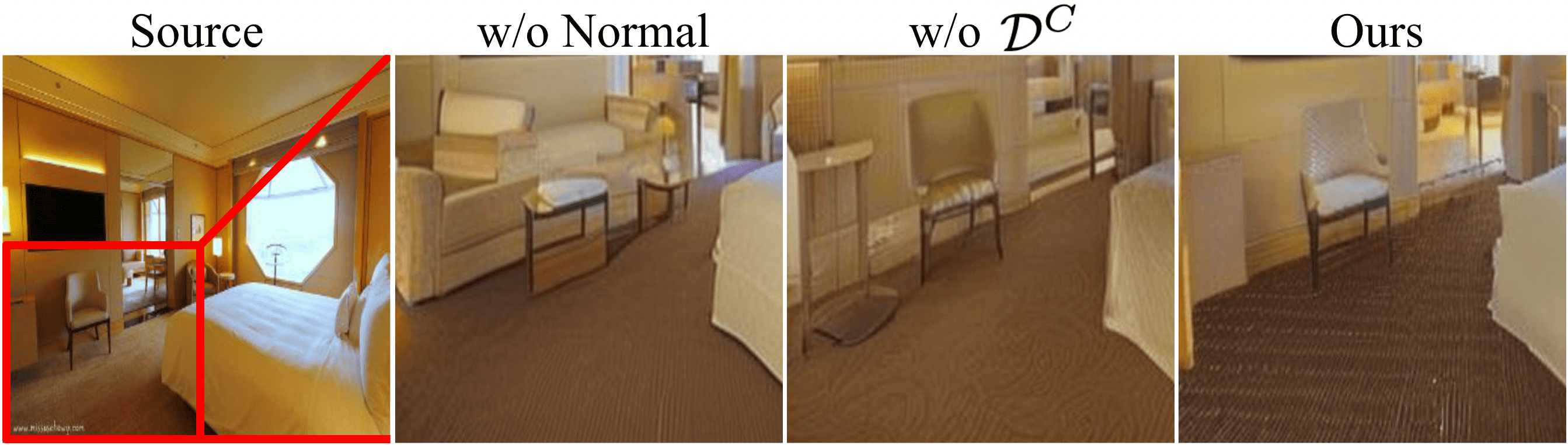} \\
    \resizebox{0.75\linewidth}{!}{%
    \begin{tabular}{cccccc}
        \toprule
         Normal & $\mathcal{D}^C$ & \bfseries{RMSE $\downarrow$} & \bfseries{PSNR $\uparrow$} & \bfseries{LPIPS $\downarrow$} \\
        \midrule
        \textendash & \checkmark & 0.2224 & 13.61 & 0.4251 \\
        \checkmark & \textendash & 0.2098 & 14.06 & 0.4093 \\
        \checkmark & \checkmark & \textbf{0.2059} & \textbf{14.29} & \textbf{0.3942} \\
        \bottomrule
    \end{tabular}
    }
    \caption{Inclusion of both normals and the control-decoder improves geometry preservation in the relit images.}
    \label{tab:ablations_control}
\end{table}

\subsection{Ablation study}\label{sec:ablation}

\noindent \textbf{Design of Albedo-conditioned Stable Image Diffusion.} 

\noindent A key contributor to the quality of the relighting is the albedo-conditioning of the diffusion model, ensuring better preservation of the intrinsic color and texture of the input. The robustness of the albedo-conditioned diffusion process is further improved by adding a noise $\epsilon^{A}_{T}$ to the latent albedo condition $z^{A}$ (in addition to the noise $\epsilon^{I}_{t}$ added to the image latent $z^{I}$). We ablate the efficacy of both components using auto-generated scribbles in two experiments: (i) we directly add the albedo as input to the ControlNet instead of conditioning Stable Diffusion on the albedo, and (ii) we train the Albedo-conditioned Stable Diffusion without adding noise to the albedo latent.  From~\Cref{tab:ablation_noisy}, we observe that albedo conditioning better preserves the intrinsic colors and texture than injecting the albedo via ControlNet (2nd row, 1st column in figure, 1st row in table vs 2nd row, 2nd column in figure, 2nd row in table). 
Furthermore, we observe that adding noise to the albedo latent (2nd row, 3rd column in the figure, 3rd row in the table) is more robust to inaccuracies in the predicted albedo (\eg, the soft glow around the lamp) and provides larger variations in lighting effects (\eg, blue sunlight peeking through the window).

\noindent \textbf{Design of ScribbleLight ControlNet.} Another key component in ScribbleLight is the control encoder-decoder for normal and scribble maps. We verify the importance of the normal map \textbf{N}, and the regularizing role of the control decoder $\mathcal{D}^C$. \Cref{tab:ablations_control} indicates that omitting the normal map (2nd column in the figure, 1st row in table) results in the creation of more random objects, even in empty spaces, due to the absence of 3D geometric guidance. Furthermore, omitting the regularizing control-decoder $\mathcal{D}^C$ (3rd column in the figure, 2nd row in table) also creates hallucinations (4th column in the figure, 3rd row in table).

\begin{figure}[h]
  \includegraphics[width=\linewidth]{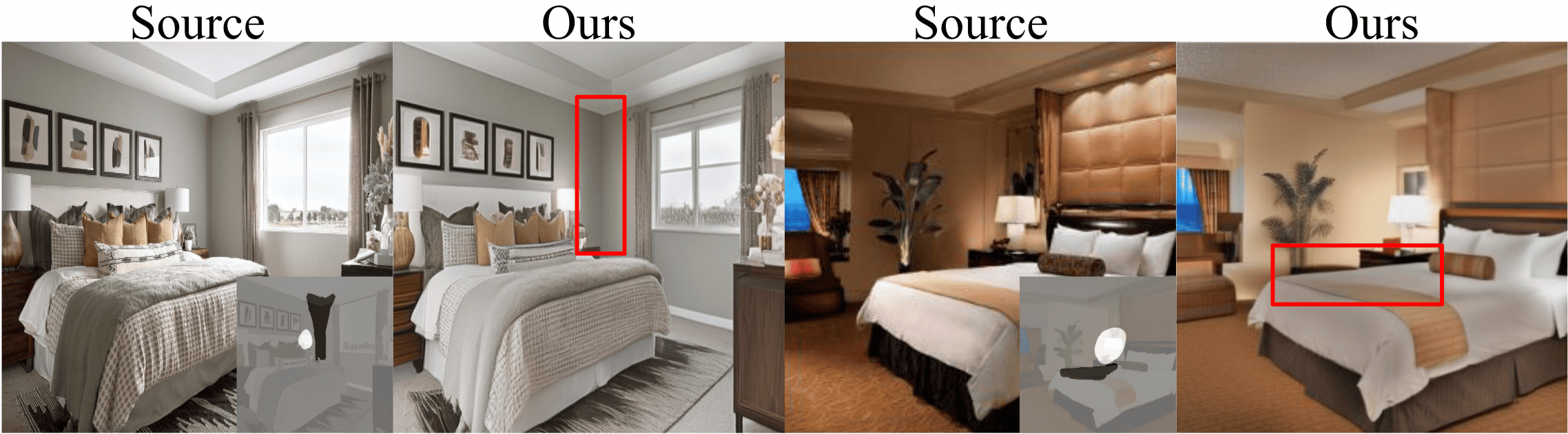}
  \caption{Given a physically incorrect scribble, \ie, the location of shadow does not match the light source, ScribbleLight often creates implausible lighting effects that best match the user scribbles.}
  \label{fig:limitation}
\end{figure}

\noindent \textbf{Limitations.} ScribbleLight struggles to rectify strong physical inconsistencies in the user-defined scribbles and often generates realistic but physically implausible lighting effects (\Cref{fig:limitation}). Furthermore, ScribbleLight does not support colored lighting adjustments, leading to relit results biased toward commonly seen colors like yellow and blue from its learned prior or source image colors.

\section{Conclusion}
In this paper, we introduce ScribbleLight, a generative model for scribble-based single-image relighting of indoor scenes. We show that scribbles are a viable control signal for realistic and physically plausible relighting while significantly reducing user efforts and providing flexibility by enabling progressive coarse-to-fine editing. Our key technical contributions are the introduction of an Albedo-conditioned Stable Image Diffusion variant that better preserves the intrinsic color and texture of the input image during relighting, and ScribbleLight's ControlNet that better preserves geometrical shading information and guides the relighting based on a latent encoding of the surface normals and rough scribble annotations. Our method outperforms existing image-based relighting algorithms adapted for scribble-based relighting in both quantitative and qualitative evaluations. We demonstrate the effectiveness of ScribbleLight in generating various lighting effects, \eg, turning a light source on or off or adding strong highlights and cast shadows.
We also show that ScribbleLight can generate multiple replicates that match the scribbles. For future work we would like to enhance scribble generation to improve user control and precision in relighting. Additional avenues for future research include incorporating colored scribbles to allow users to control the color of the lighting.

{
    \small
    \bibliographystyle{ieeenat_fullname}
    \bibliography{main}
}

\clearpage
\setcounter{page}{1}
\maketitlesupplementary
\appendix

\section{Implementation Details} \label{sec:implementation}
We fine-tune a pre-trained Stable Diffusion v2 model~\cite{rombach2022stablediffusion} for Albedo-conditioned Image Diffusion and ScribbleLight ControlNet. To reconstruct the monochromatic shading map $\textbf{S}_{mono}$ and the normal map \textbf{N} from the lighting feature map $f$, we utilize a control decoder $\mathcal{D}^C$. This control decoder $\mathcal{D}^C$ is structured similarly to the control encoder $\mathcal{E}^C$, consisting of 4 residual blocks, but with a transposed architecture. For training, we use a batch size of 16 and the AdamW optimizer with a learning rate of 1e\textendash5. All inputs are resized to 512 × 512. Training each model takes approximately 48 hours on 4 A6000 GPUs. We employ the DDPM noise scheduler with 1000 diffusion steps during training. For inference, we apply the DDIM scheduler and sample only 20 steps.

\section{Evaluation with Monochromatic Shading} \label{sec:mono}
This section is analogous to \cref{sec:experiment}, but instead of using scribbles, we evaluate indoor scene relighting performance using the monochromatic shading map $\textbf{S}_{mono}$.

\noindent \textbf{Dataset and Baselines.}
As detailed in \cref{sec:eval_framework}, we trained both LightIt*~\cite{kocsis2024lightit} and our method using the LSUN bedrooms dataset~\cite{yu2015lsun}. Instead of using auto-generated scribbles as the ControlNet input, however, here we utilize the monochromatic shading map instead. For our second baseline, we employed RGB$\leftrightarrow$X~\cite{zeng2024rgbx} without retraining: intrinsic components (normal, albedo, roughness, and metallicity) were extracted from the source image, and the irradiance field was derived from the target image using RGB$\rightarrow$X. The source image was then relit by recomposing it with its intrinsic components and the target image's irradiance field through X$\rightarrow$RGB.
Since IIDiffusion~\cite{kocsis2024iidiffusion} employs spherical Gaussians as its lighting representation, it was not feasible to perform a comparison based on scribbles. However, we extend the comparison with IIDiffusion by extracting intrinsic components from the source image and the spherical Gaussians from the target image, and recomposing the source image under the target spherical Gaussians, similar to RGB$\leftrightarrow$X.

\noindent \textbf{Evaluation.}
We quantitatively compare the relighting quality using the monochromatic shading $\textbf{S}_{mono}$ over the test set. The results, summarized in~\cref{tab:sup}, show that ScribbleLight outperforms the baseline methods across all metrics. Additionally, we present the qualitative results on the webpage (see the `Monochromatic Shading Map' section).

\begin{table}[t]
  \centering
  \resizebox{\linewidth}{!}{%
  \begin{tabular}{lcccc}
    \toprule
     & {\bfseries{RMSE} $\downarrow$} & {\bfseries{PSNR} $\uparrow$} & {\bfseries SSIM $\uparrow$} & {\bfseries LPIPS $\downarrow$} \\
    \midrule
    IIDiffusion 
    & 0.137 & 18.02 & 0.690 & 0.367  \\
    LightIt* 
    & 0.227 & 13.17 & 0.390 & 0.447  \\
    RGB$\leftrightarrow$X 
    & 0.261 & 13.36 & 0.570 & 0.364  \\
    Ours  
    & \textbf{0.132} & \textbf{18.22} & \textbf{0.697} & \textbf{0.275}  \\
  \bottomrule
\end{tabular}%
}
\caption{Quantitative comparison of relighting accuracy between our ScribbleLight, RGB$\leftrightarrow$X~\cite{zeng2024rgbx}, LightIt*~\cite{kocsis2024lightit}, and IIDiffusion~\cite{kocsis2024iidiffusion} with monochromatic shading map created from the target image. We compute the errors with respect to a target image.
}
\label{tab:sup}
\end{table}

\section{Webpage Materials}
\begin{figure*}[h]
  \includegraphics[width=\linewidth]{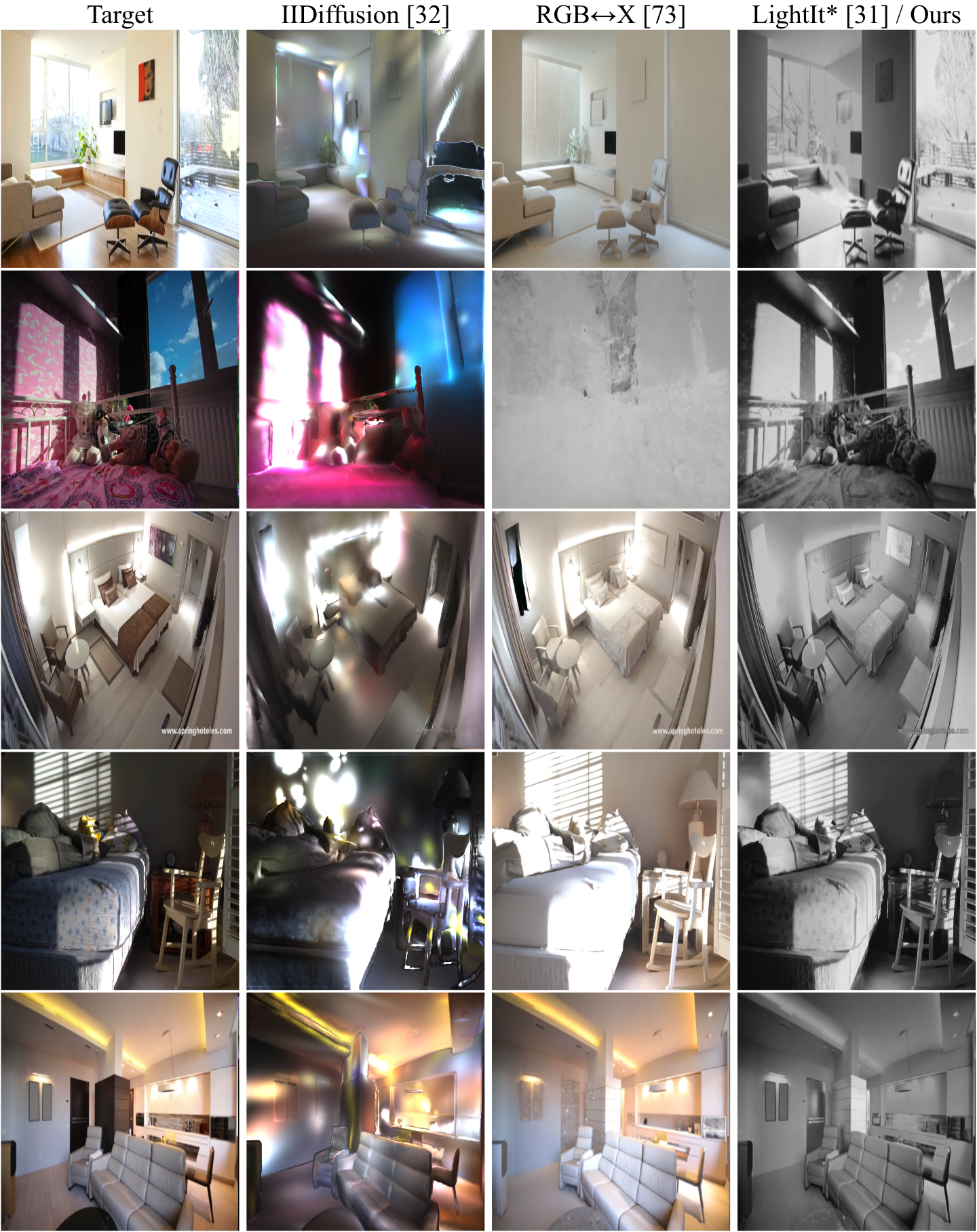}
  \caption{We demonstrate the target lighting representation used by each method—IIDiffusion~\cite{kocsis2024iidiffusion}, RGB$\leftrightarrow$X~\cite{zeng2024rgbx}, LightIt*~\cite{kocsis2024lightit}, and Ours—when performing relighting with a monochromatic shading map.}
  \label{fig:suppl}
\end{figure*}

The structure of our webpage is as follows. As demonstrated in \cref{fig:teaser} and \cref{fig:iterative}, the top section presents a demo video and iterative examples showcasing how ScribbleLight iteratively refines lighting effects. The `User Scribble' section provides additional examples of user scribble comparisons, as illustrated in \cref{fig:user_scrib}. The `Monochromatic Shading Map' section features qualitative comparisons referenced in \cref{sec:mono}. We also present the target shading utilized by each method during relighting in \cref{fig:suppl}. Finally, the `Turning On/Off the Light' section includes additional examples from \cref{fig:on_off}.

\end{document}